\documentclass{article}

\PassOptionsToPackage{square,sort,comma,numbers}{natbib}
\usepackage{iclr2017_conference,times}

\usepackage[utf8]{inputenc} 
\usepackage[T1]{fontenc}    
\usepackage{url}            
\usepackage{booktabs}       
\usepackage{amsfonts}       
\usepackage{amsmath}        
\usepackage{nicefrac}       
\usepackage{graphicx}       
\usepackage{microtype}      
\usepackage{subfig}      
\usepackage{amsmath}
\usepackage{soul} 
\usepackage[font={footnotesize}]{caption}
\usepackage{enumitem}
\usepackage[normalem]{ulem}
\usepackage{rotating}
\usepackage{array}

\usepackage{floatrow}
\newfloatcommand{capbtabbox}{table}[][\FBwidth]
\setlist{nosep}

\usepackage{color}
\definecolor{darkGreen}{rgb}{0, 0.5, 0.2}
\definecolor{orange}{rgb}{1, 0.75, 0.5}
\definecolor{purple}{rgb}{0.5, 0.2, 0.9}

\title{Learning to reinforcement learn}

\author{\small{JX Wang$^1$, \ Z Kurth-Nelson$^1$, \ D Tirumala$^1$, \ H Soyer$^1$, \ JZ Leibo$^1$,} 
\vspace{0.1 cm}
\\\textbf{\small{R Munos$^1$, \ C Blundell$^1$, \ D Kumaran$^{1,3}$, \ M Botvinick$^{1,2}$}}
\vspace{0.1 cm}
\\
$^1$DeepMind, London, UK \\
$^2$Gatsby Computational Neuroscience Unit, UCL, London, UK \\
$^3$Institute of Cognitive Neuroscience, UCL, London, UK \\
\vspace{0.15 cm}
\\
\scriptsize\texttt{\{wangjane, zebk, dhruvat, soyer, jzl, munos, cblundell,} \\
\scriptsize\texttt{dkumaran, botvinick\} @google.com} \\
}
\begin{document}

\maketitle

\begin{abstract}

In recent years deep reinforcement learning (RL) systems have attained superhuman performance in a number of challenging task domains. However, a major limitation of such applications is their demand for massive amounts of training data. A critical present objective is thus to develop deep RL methods that can adapt rapidly to new tasks. In the present work we introduce a novel approach to this challenge, which we refer to as deep meta-reinforcement learning. Previous work has shown that recurrent networks can support meta-learning in a fully supervised context. We extend this approach to the RL setting. What emerges is a system that is trained using one RL algorithm, but whose recurrent dynamics implement a second, quite separate RL procedure. This second, learned RL algorithm can differ from the original one in arbitrary ways. Importantly, because it is learned, it is configured to exploit structure in the training domain. We unpack these points in a series of seven proof-of-concept experiments, each of which examines a key aspect of deep meta-RL. We consider prospects for extending and scaling up the approach, and also point out some potentially important implications for neuroscience.

\end{abstract}

\section{Introduction}

Recent advances have allowed long-standing methods for reinforcement learning (RL) to be newly extended to such complex and large-scale task environments as Atari \citep{Mnih2015} and Go \citep{silver2016mastering}. The key enabling breakthrough has been the development of techniques allowing the stable integration of RL with non-linear function approximation through deep learning \citep{Mnih2015,lecun2015deep}. The resulting deep RL methods are attaining human- and often superhuman-level performance in an expanding list of domains \citep{Mnih2015,silver2016mastering,JaderbergDreamingICLR}. However, there are at least two aspects of human performance that they starkly lack. First, deep RL typically requires a massive volume of training data, whereas human learners can attain reasonable performance on any of a wide range of tasks with comparatively little experience. Second, deep RL systems typically specialize on one restricted task domain, whereas human learners can flexibly adapt to changing task conditions. Recent critiques \citep[e.g.,][]{lake2016building} have invoked these differences as posing a direct challenge to current deep RL research.
 
In the present work, we outline a framework for meeting these challenges, which we refer to as \textit{deep meta-reinforcement learning}, a label that is intended to both link it with and distinguish it from previous work employing the term ``meta-reinforcement learning'' \citep[e.g.][discussed later]{schmidhuber1996simple, schweighofer2003meta}. The key concept is to use standard deep RL techniques to train a recurrent neural network in such a way that the recurrent network comes to implement its own, free-standing RL procedure. As we shall illustrate, under the right circumstances, the secondary learned RL procedure can display an adaptiveness and sample efficiency that the original RL procedure lacks.

The following sections review previous work employing recurrent neural networks in the context of meta-learning and describe the general approach for extending such methods to the RL setting.  We then present seven proof-of-concept experiments, each of which highlights an important ramification of the deep meta-RL setup by characterizing agent performance in light of this framework. We close with a discussion of key challenges for next-step research, as well as some potential implications for neuroscience.

\section{Methods}
\subsection{Background: Meta-learning in recurrent neural networks}

Flexible, data-efficient learning naturally requires the operation of prior biases.  In general terms, such biases can derive from two sources; they can either be engineered into the learning system (as, for example, in convolutional networks), or they can themselves be acquired through learning. The second case has been explored in the machine learning literature under the rubric of \textit{meta-learning} \citep{thrun1998learning,schmidhuber1996simple}.
 
In one standard setup, the learning agent is confronted with a series of tasks that differ from one another but also share some underlying set of regularities. Meta-learning is then defined as an effect whereby the agent improves its performance in each new task more rapidly, on average, than in past tasks \citep{thrun1998learning}. At an architectural level, meta-learning has generally been conceptualized as involving two learning systems: one lower-level system that learns relatively quickly, and which is primarily responsible for adapting to each new task; and a slower higher-level system that works across tasks to tune and improve the lower-level system.
 
A variety of methods have been pursued to implement this basic meta-learning setup, both within the deep learning community and beyond \citep{thrun1998learning}. Of particular relevance here is an approach introduced by Hochreiter and colleagues \citep{hochreiter2001learning}, in which a recurrent neural network is trained on a series of interrelated tasks using standard backpropagation. A critical aspect of their setup is that the network receives, on each step within a task, an auxiliary input indicating the target output for the preceding step. For example, in a regression task, on each step the network receives as input an \textit{x} value for which it is desired to output the corresponding \textit{y}, but the network also receives an input disclosing the target \textit{y} value for the preceding step \citep[see][]{hochreiter2001learning,santoro2016meta}. In this scenario, a different function is used to generate the data in each training episode, but if the functions are all drawn from a single parametric family, then the system gradually tunes into this consistent structure, converging on accurate outputs more and more rapidly across episodes.
 
One interesting aspect of Hochreiter’s method is that the process that underlies learning within each new task inheres entirely in the dynamics of the recurrent network, rather than in the backpropagation procedure used to tune that network’s weights. Indeed, after an initial training period, the network can improve its performance on new tasks even if the weights are held constant \citep[see also ][]{prokhorov2002adaptive,cotter1990fixed,younger1999fixed}. A second important aspect of the approach is that the learning procedure implemented in the recurrent network is fit to the structure that spans the family of tasks on which the network is trained, embedding biases that allow it to learn efficiently when dealing with tasks from that family.

\subsection{Deep meta-RL: Definition and key features}

Importantly, Hochreiter’s original work \citep{hochreiter2001learning}, as well as its subsequent extensions \citep{santoro2016meta,prokhorov2002adaptive,cotter1990fixed,younger1999fixed} only addressed supervised learning (i.e. the auxiliary input provided on each step explicitly indicated the target output on the previous step, and the network was trained using explicit targets). In the present work we consider the implications of applying the same approach in the context of reinforcement learning. Here, the tasks that make up the training series are interrelated RL problems, for example, a series of bandit problems varying only in their parameterization.  Rather than presenting target outputs as auxiliary inputs, the agent receives inputs indicating the action output on the previous step and, critically, the quantity of reward resulting from that action. The same reward information is fed in parallel to a deep RL procedure, which tunes the weights of the recurrent network.
 
It is this setup, as well as its result, that we refer to as deep meta-RL (although from here on, for brevity, we will often simply call it meta-RL, with apologies to authors who have used that term previously). As in the supervised case, when the approach is successful, the dynamics of the recurrent network come to implement a learning algorithm entirely separate from the one used to train the network weights. Once again, after sufficient training, learning can occur within each task even if the weights are held constant. However, here the procedure the recurrent network implements is itself a full-fledged reinforcement learning algorithm, which negotiates the exploration-exploitation tradeoff and improves the agent’s policy based on reward outcomes. A key point, which we will emphasize in what follows, is that this learned RL procedure can differ starkly from the algorithm used to train the network’s weights. In particular, its policy update procedure (including features such as the effective learning rate of that procedure), can differ dramatically from those involved in tuning the network weights, and the learned RL procedure can implement its own approach to exploration. Critically, as in the supervised case, the learned RL procedure will be fit to the statistics spanning the multi-task environment, allowing it to adapt rapidly to new task instances.

\subsection{Formalism}\label{sec:formalism}
Let us write as ${\cal D}$ a distribution (the prior) over Markov Decision Processes (MDPs). We want to demonstrate that meta-RL is able to learn a prior-dependent RL algorithm, in the sense that it will perform well on average on MDPs drawn from ${\cal D}$ or slight modifications of ${\cal D}$. An appropriately structured agent, embedding a recurrent neural network, is trained by interacting with a sequence of MDP environments (also called tasks) through episodes. At the start of a new episode, a new MDP task $m\sim {\cal D}$ and an initial state for this task are sampled, and the internal state of the agent (i.e., the pattern of activation over its recurrent units) is reset.  The agent then executes its action-selection strategy in this environment for a certain number of discrete time-steps. At each step $t$ an action $a_t\in A$ is executed as a function of the whole history ${\cal H}_{t}=\{ x_0, a_0, r_0, \dots, x_{t-1}, a_{t-1}, r_{t-1}, x_t\} $ of the agent interacting in the MDP $m$ during the current episode (set of states $\{x_s\}_{0\leq s\leq t}$, actions $\{ a_s\}_{0\leq s<t}$, and rewards $\{ r_s\}_{0\leq s<t}$ observed since the beginning of the episode, when the recurrent unit was reset). The network weights are trained to maximize the sum of observed rewards over all steps and episodes.

After training, the agent’s policy is fixed (i.e. the weights are frozen, but the activations are changing due to input from the environment and the hidden state of the recurrent layer), and it is evaluated on a set of MDPs that are drawn either from the same distribution ${\cal D}$ or slight modifications of that distribution (to test the generalization capacity of the agent). The internal state is reset at the beginning of the evaluation of any new episode. Since the policy learned by the agent is history-dependent (as it makes uses of a recurrent network), when exposed to any new MDP environment, it is able to adapt and deploy a strategy that optimizes rewards for that task.

\section{Experiments}

In order to evaluate the approach to learning that we have just described, we conducted a series of six proof-of-concept experiments, which we present here along with a seventh experiment originally reported in a related paper \citep{MirowskiICLR17}. One particular point of interest in these experiments was to see whether meta-RL could be used to learn an adaptive balance between exploration and exploitation, as demanded of any fully-fledged RL procedure. A second and still more important focus was on the question of whether meta-RL can give rise to learning that gains efficiency by capitalizing on task structure. 

\begin{figure}[ht]
  \centering
    \includegraphics[width=0.65\textwidth,natwidth=610,natheight=642]{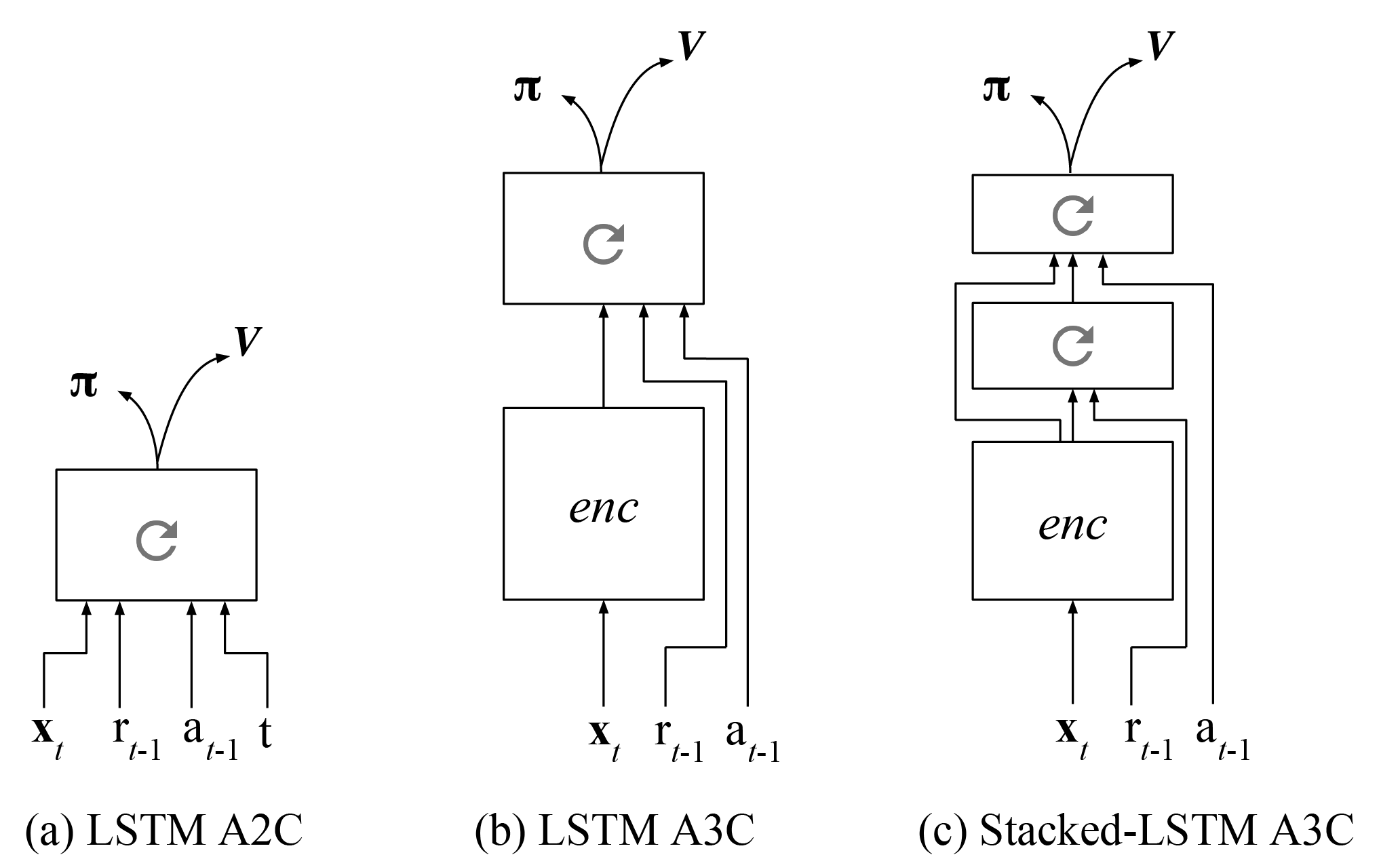}
    
    \caption{ Advantage actor-critic with recurrence. In all architectures, reward and last action are additional inputs to the LSTM. For non-bandit environments, observation is also fed into the LSTM either as a one-hot or passed through an encoder model [3-layer encoder: two convolutional layers (first layer: 16 8x8 filters applied with stride 4, second layer: 32 4x4 filters with stride 2) followed by a fully connected layer with 256 units and then a ReLU non-linearity. See for details \cite{MirowskiICLR17}]. For bandit experiments, current time step is also fed in as input.  $\pi$ = policy; $v$ = value function.  A3C is the distributed multi-threaded asynchronous version of the advantage actor-critic algorithm \citep{mnih2016a3c}; A2C is single threaded. (a) Architecture used in experiments 1-5. (b) Convolutional-LSTM architecture used in experiment 6. (c) Stacked LSTM architecture with convolutional encoder used in experiments 6 and 7.
    }
    \label{fig:archs}
\end{figure}

In order to examine these questions, we performed four experiments focusing on bandit tasks and two additional experiments focusing on Markov decision problems. All of our experiments (as well as the additional experiment we report) employ a common set of methods, with minor implementational variations. In all experiments, the agent architecture centers on a recurrent neural network \citep[LSTM;][]{hochreiter1997long} feeding into a soft-max output representing discrete actions. As detailed below, the parameters of this network core, as well as some other architectural details, varied across experiments (see Figure \ref{fig:archs} and Table \ref{table:params}). However, it is important to emphasize that comparisons between specific architectures are outside the scope of this paper. Our main aim is to illustrate and validate the  meta-RL framework in a more general way. To this end, all experiments used the high-level task setup previously described: Both training and testing were organized into fixed-length episodes, each involving a task randomly sampled from a predetermined task distribution, with the LSTM hidden state initialized at the beginning of each episode. Task-specific inputs and action outputs are described in conjunction with individual experiments. In all experiments except where specified, the input included a scalar indicating the reward received on the preceding time-step as well as a one-hot representation of the action sampled on that time-step. 

All reinforcement learning was conducted using the Advantage Actor-Critic algorithm, as detailed in \cite{mnih2016a3c} and \cite{MirowskiICLR17} (see also Figure \ref{fig:archs}). Details of training, including the use of entropy regularization and a combined policy and value estimate loss, closely follow the methods detailed in \cite{MirowskiICLR17}, with the exception that our experiments used a single thread unless otherwise noted. For a full listing of parameters refer to Table \ref{table:params}.

\begin{table}
\caption{List of hyperparameters. $\beta_e$ = coefficient of entropy regularization loss; in Exps. 1-4, $\beta_e$ is annealed from 1.0 to 0.0 over the course of training. $\beta_v$ = coefficient of value function loss \citep{MirowskiICLR17}. $r$ = reward, $a$ = last action, $t$ = current time step, $x$ = current observation. Exp. 1: Bandits with independent arms (Section 3.1.1); Exp. 2: Bandits with dependent arms I (Section 3.1.2); Exp. 3: Bandits with dependent arms II (Section 3.1.3); Exp. 4: Restless bandits (Section 3.1.4); Exp. 5: The ``Two-Step Task'' (Section 3.2.1); Exp. 6: Learning abstract task structure (Section 3.2.2).}
\begin{center}
\begin{tabular}{l c c c c c c}
{\bf Parameter}  & {\bf Exps. 1 \& 2} & {\bf Exp. 3} & {\bf Exp. 4} & {\bf Exp. 5} & {\bf Exp. 6}\\
\hline
No. threads & 1 & 1 & 1 & 1 & 32 \\
No. LSTMs & 1 & 1 & 1 & 1 & 2 \\
No. hiddens & 48 & 48 & 48 & 48 & 256/64 \\
Steps unrolled & 100 & 5 & 150 & 20 & 100 \\
$\beta_e$ & \textit{annealed} & \textit{annealed} & \textit{annealed} & 0.05 & 0.001 \\
$\beta_v$ & 0.05 & 0.05 & 0.05 & 0.05 & 0.4 \\
Learning rate & \textit{tuned} & 0.001 & 0.001 & \textit{tuned} & \textit{tuned} \\
Discount factor & \textit{tuned} & 0.8 & 0.8 & \textit{tuned} & \textit{tuned} \\
Input & $a$, $r$, $t$ & $a$, $r$, $t$ & $a$, $r$, $t$ & $a$, $r$, $t$, $x$ & $a$, $r$, $x$ \\
Observation & n/a & n/a & n/a & 1-hot & RGB (84x84) \\
No. trials/episode & 100 & 5 & 150 & 10 & 10 \\
Episode length & 100 & 5 & 150 & 20 & $\leq$3600 \\
\hline
\end{tabular}
\end{center}
\label{table:params}
\end{table}

\subsection{Bandit problems}

As an initial setting for evaluating meta-RL, we studied a series of bandit problems. Except for a very limited set of bandit environments, it is intractable to compute the (prior-dependent) Bayesian-optimal strategy. Here we demonstrate that a recurrent system trained on a set of bandit environments drawn i.i.d.~from a given distribution of environments produces a bandit algorithm which performs well on problems drawn from that distribution, and to a certain extent generalizes to related distributions. Thus, meta-RL learns a prior-dependent bandit algorithm. 

The specific bandit instantiation of the general meta-RL procedure described in Section~\ref{sec:formalism} is defined as follows. Let ${\cal D}$ be a {\em training distribution} over bandit environments. The meta-RL system is trained on a sequence of bandit environments through episodes. At the start of a new episode, its LSTM state is reset and a bandit task $b\sim {\cal D}$ is sampled. A bandit task is defined as a set of distributions -- one for each arm -- from which rewards are sampled. The agent plays in this bandit environment for a certain number of trials and is trained to maximize observed rewards. After training, the agent's policy is evaluated on a set of bandit tasks that are drawn from a {\em test distribution}  ${\cal D'}$, which can either be the same as ${\cal D}$ or a slight modification of it.

We evaluate the resulting performance of the learned bandit algorithm by the cumulative regret, a measure of the loss (in expected rewards) suffered when playing sub-optimal arms. Writing $\mu_a(b)$ the expected reward of arm $a$ in bandit environment $b$, and $\mu^*(b)=\max_a \mu_a(b)=\mu_{a^*(b)}(b)$ (where $a^*(b)$ is one optimal arm) the optimal expected reward, we define the cumulative regret (in environment $b$) as $R_T(b) = \sum_{t=1}^T \mu^*(b) - \mu_{a_t}(b)$, where $a_t$ is the arm (action) chosen at time $t$. In experiment 4 (Restless bandits; Section 3.1.4), $\mu^*$ also depends on $t$. We report the performance (average over bandit environments drawn from the test distribution) either in terms of the cumulative regret: ${\mathbb E}_{b\sim {\cal D'}}[R_T(b)]$ or in terms of number of sub-optimal pulls:  ${\mathbb E}_{b\sim {\cal D'}}[\sum_{t=1}^T {\mathbb I}\{a_t \neq a^*(b)\} ]$.

\subsubsection{Bandits with independent arms}
We first consider a simple two-armed bandit task to examine the behavior of meta-RL under conditions where theoretical guarantees exist and general purpose algorithms apply. The arm distributions are independent Bernoulli distributions (rewards are $1$ with probability $p$ and $0$ with probability $1-p$), where the parameters of each arm ($p_1$ and $p_2$) are sampled independently and uniformly over $[0,1]$. We denote by ${\cal D}_i$ the corresponding distribution over these independent bandit environments (where the subscript $i$ stands for independent arms). 

At the beginning of each episode, a new bandit task is sampled and held constant for 100 trials. Training lasted for 20,000 episodes.
The network is given as input the last reward, last action taken, and the trial number $t$, subsequently producing the action for the next trial $t+1$ (Figure \ref{fig:archs}). After training, we evaluated on 300 new episodes with the learning rate set to zero (the learned policy is fixed). 

Across model instances, we randomly sampled learning rate and discount, following \cite{mnih2016a3c}.
For all figures, we plotted the average of the top 5 runs of 100 randomly sampled hyperparameter settings, where the top agents were selected from the first half of the 300 evaluation episodes and performance was plotted for the second half. We measured the cumulative expected regret across the episode, comparing with several algorithms tailored for this independent bandit setting: Gittins indices \citep{gittins1979bandit} (which is Bayesian optimal in the finite-horizon case), UCB \citep{auer2002finite} (which comes with theoretical finite-time regret guarantees), and Thompson sampling~\citep{thompson1933likelihood} \citep[which is asymptotically optimal in this setting: see][]{kaufmann2012}. Model simulations were conducted with the PymaBandits toolbox from \citep{kaufmann2012pyma} and custom Matlab scripts.

\begin{figure}
    \centering
        \includegraphics[width=0.85\textwidth,natwidth=610,natheight=642]{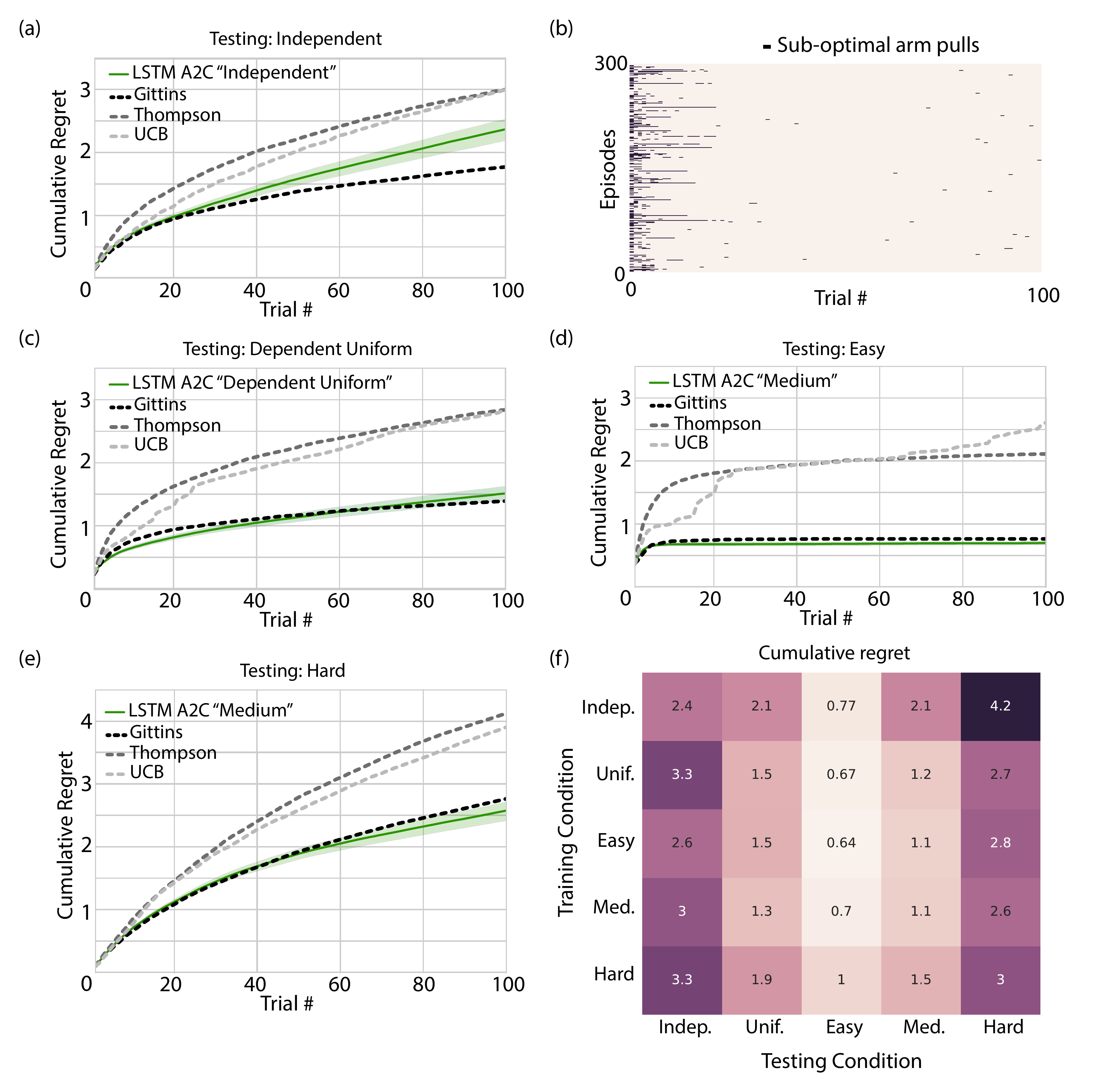}
    \caption{Performance on independent- and correlated-arm bandits. We report performance as the cumulative expected regret $R_T$ for 150 test episodes, averaged over the top 5 hyperparameters for each agent-task configuration, where the top 5 was determined based on performance on a separate set of 150 test episodes. (a) LSTM A2C trained and evaluated on bandits with independent arms (distribution ${\cal D}_i$; see text), and compared with theoretically optimal models. (b) A single agent playing the medium difficulty task with distribution ${\cal D}_m$. Suboptimal arm pulls over trials are depicted for 300 episodes. (c) LSTM A2C trained and evaluated on bandits with dependent uniform arms (distribution ${\cal D}_u$), (d) trained on medium bandit tasks (${\cal D}_m$) and tested on easy (${\cal D}_e$), and (e) trained on medium (${\cal D}_m$) and tested on hard task (${\cal D}_h$). (f) Cumulative regret for all possible combinations of training and testing environments (${\cal D}_i$, ${\cal D}_u$, ${\cal D}_e$, ${\cal D}_m$, ${\cal D}_h$).}
    \label{fig:simplebandits}
\end{figure}

As shown in Figure \ref{fig:simplebandits}a (green line; ``Independent''), meta-RL outperforms both Thompson sampling (gray dashed line) and UCB (light gray dashed line), although it performs less well compared to Gittins (black dashed line). To verify the critical importance of providing reward information to the LSTM, we removed this input, leaving all other inputs as before. As expected, performance was at chance levels on all bandit tasks.

\subsubsection{Bandits with dependent arms (I)}

As we have emphasized, a key property of meta-RL is that it gives rise to a learned RL algorithm that exploits consistent structure in the training distribution. In order to garner empirical evidence for this point, we tested the agent from our first experiment in a more structured bandit task. Specifically, we trained the system on two-arm bandits in which arm reward distributions are correlated. In this setting, unlike the one studied in the previous section, experience with either arm provides information about the other. Standard bandit algorithms, including UCB and Thompson sampling, perform suboptimally in this setting, as they are not designed to exploit such correlations. In some cases it is possible to tailor algorithms for specific arm structures \citep[see for example][]{structured_bandits2014}, but extensive problem-specific analysis is typically required. Our approach aims to learn a structure-dependent bandit algorithm directly from experience with the target bandit domain. 

We consider Bernoulli distributions where the parameters $(p_1,p_2)$ of the two arms are correlated in the sense that $p_1=1-p_2$. We consider several training and test distributions. The {\em uniform} means that $p_1\sim {\cal U}([0,1])$ (uniform distribution over the unit interval). The {\em easy} means that $p_1\sim {\cal U}(\{0.1,0.9\})$ (uniform distribution over those two possible values), and similarly we call {\em medium} when $p_1\sim {\cal U}(\{0.25,0.75\})$ and {\em hard} when $p_1\sim {\cal U}(\{0.4,0.6\})$. We denote by ${\cal D}_u$, ${\cal D}_e$, ${\cal D}_m$, and ${\cal D}_h$ the corresponding induced distributions over bandit environments. In addition we also considered the independent uniform distribution (as in the previous section, ${\cal D}_i$) where $p_1,p_2\sim {\cal U}([0,1])$ independently. Agents were both trained and tested on those five distributions over bandit environments (among which four correspond to correlated distributions: ${\cal D}_u$, ${\cal D}_e$, ${\cal D}_m$ and ${\cal D}_h$; and one to the independent case: ${\cal D}_i$). As a validation of the names given to the task distributions (${\cal D}_e$, ${\cal D}_m$, ${\cal D}_h$), results show that the easy task is easier to learn than the medium which itself is easier than the hard one (Figure \ref{fig:simplebandits}f). This is compatible with the general notion that the hardness of a bandit problem is inversely proportional to the difference between the expected reward of the optimal and sub-optimal arms. We again note that withholding the reward input to the LSTM resulted in chance performance on even the easiest bandit task, as should be expected. 

Figure \ref{fig:simplebandits}f reports the results of all possible training-testing regimes. From observing the cumulative expected regrets, we make the following observations: i) agents trained in structured environments (${\cal D}_u$, ${\cal D}_e$, ${\cal D}_m$, and ${\cal D}_h$) develop prior knowledge that can be used effectively when tested on structured distributions -- performing comparably to Gittins (Figure \ref{fig:simplebandits}c-f), and superiorly compared to agents trained on independent arms (${\cal D}_i$) in all structured tasks at test (Figure \ref{fig:simplebandits}f). This is because an agent trained on independent rewards (${\cal D}_i$) has not learned to exploit the reward correlations that are useful in those structured tasks. ii) Conversely, previous training on any structured distribution (${\cal D}_u$, ${\cal D}_e$, ${\cal D}_m$, or ${\cal D}_h$) hurts performance when agents are tested on an independent distribution (${\cal D}_i$; Figure \ref{fig:simplebandits}f). This makes sense, as training on correlated arms may produce a policy that relies on specific reward structure, thereby impacting performance in problems where no such structure exists. iii) Whilst the previous results emphasize the point that meta-RL gives rise to a separate learnt RL algorithm that implements prior-dependent bandit strategies, results also provide evidence that there is some generalization beyond the exact training distribution encountered (Figure \ref{fig:simplebandits}f). For example, agents trained on the distributions ${\cal D}_e$ and ${\cal D}_m$ perform well when tested over a much wider structured distribution (i.e. ${\cal D}_u$). Further, our evidence suggests that there is generalization from training on the easier tasks (${\cal D}_e$,${\cal D}_m$) to testing on the hardest task (${\cal D}_h$; Figure \ref{fig:simplebandits}e), with similar or even marginally superior performance as compared to training on the hard distribution ${\cal D}_h$ itself(Figure \ref{fig:simplebandits}f). In contrast, training on the hard distribution ${\cal D}_h$ results in relatively poor generalization to other structured distributions (${\cal D}_u$, ${\cal D}_e$, ${\cal D}_m$), suggesting that training purely on hard instances may result in a learned RL algorithm that is more constrained by prior knowledge, perhaps due to the difficulty of solving the original problem. 

\subsubsection{Bandits with dependent arms (II)}

In the previous experiment, the agent could outperform standard bandit algorithms by making use of learned dependencies between arms. However, it could do this while always choosing what it believes to be the highest-paying arm. We next examine a problem where information can be gained by paying a short-term reward cost. Similar problems have been examined before as providing a challenge to standard bandit algorithms \citep[see e.g.][]{Russo2014}. In contrast, humans and animals make decisions that sacrifice immediate reward for information gain \citep[e.g.][]{bromberg2009midbrain}.

In this experiment, the agent was trained on 11-armed bandits with strong dependencies between arms. All arms had deterministic payouts. Nine ``non-target'' arms had reward $= 1$, and one ``target'' arm had reward $=5$. Meanwhile, arm $a_{11}$ was always ``informative'', in that the target arm was indexed by 10 times $a_{11}$'s reward (e.g. a reward of 0.2 on $a_{11}$ indicated that $a_2$ was the target arm). Thus, $a_{11}$'s payouts ranged from 0.1 to 1. In each episode, the index of the target arm was randomly assigned. On the first trial of each episode, the agent could not know which arm was the target, so the informative arm returned expected reward 0.55 and every target arm returned expected reward 1.4. Choosing the informative arm thus meant foregoing immediate reward, but with the compensation of valuable information. Episodes were five steps long. Again, the reward on the previous trial was provided as an additional observation to the agent. To facilitate learning, this was encoded in 1-hot format.

\begin{figure}
    \centering
    \includegraphics[width=0.4\textwidth,natwidth=610,natheight=642]{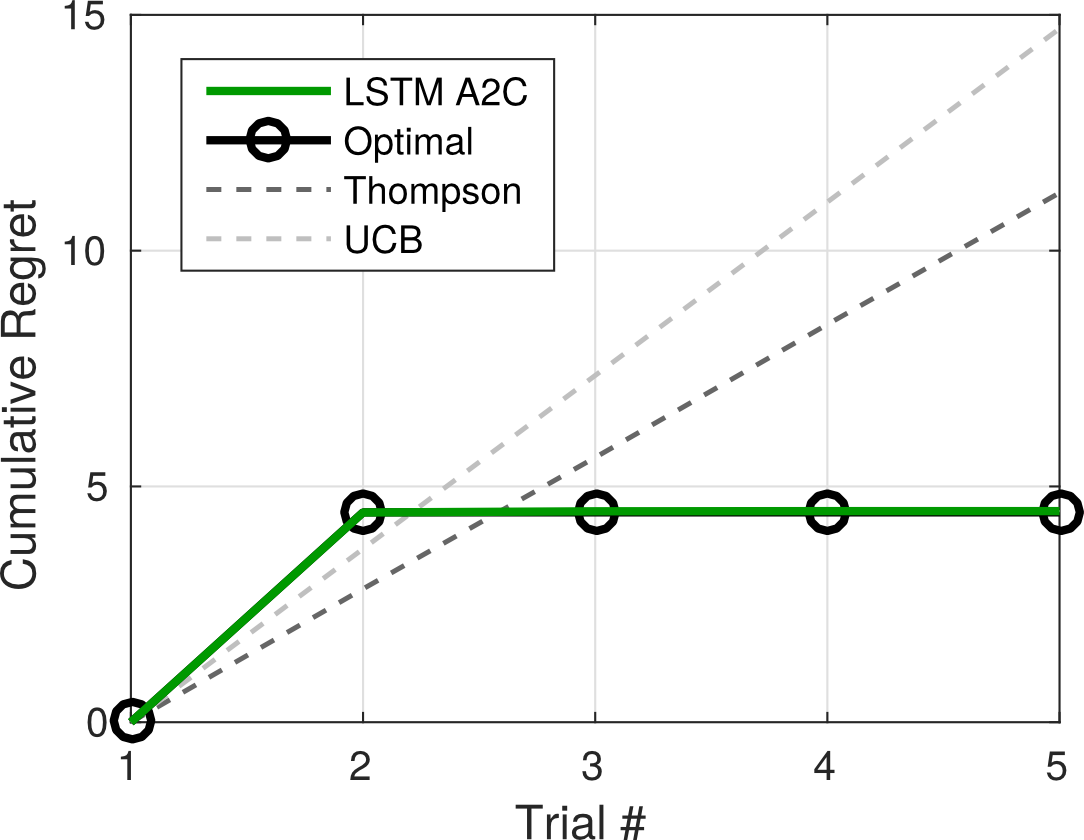}
    \caption{Learned RL procedure pays immediate cost to gain information to improve long-run returns. In this task, one arm is lower-paying but provides perfect information about which of the other ten arms is highest-paying. The remaining nine arms are intermediate in reward. The index of the informative arm is fixed between episodes, but the index of the highest-paying arm is randomized between episodes. On the first trial, the trained agent samples the informative arm. On subsequent trials, the agent uses the information it gained to deterministically exploit the highest-paying arm. Thompson sampling and UCB are not able to take advantage of the dependencies between arms. }
    \label{fig:informbandits}
\end{figure}

Results are shown in Figure \ref{fig:informbandits}. The agent learned the optimal long-run strategy of sampling the informative arm once, despite the short-term cost, and then using the resulting information to exploit the high-value target arm. Thompson sampling, if supplied the true prior, searched potential target arms and exploited the target if found. UCB performed worse because it sampled every arm once even if the target arm was found early.

\subsubsection{Restless bandits}
In previous experiments we considered stationary problems where the agent's actions yielded information about task parameters that remained fixed throughout each episode. Next, we consider a bandit problem in which reward probabilities change over the course of an episode, with different rates of change (volatilities) in different episodes. To perform well, the agent must not only track the best arm, but also infer the volatility of the episode and adjust its own learning rate accordingly. In such an environment, learning rates should be higher when the environment is changing rapidly, because past information becomes irrelevant more quickly \citep{sutton1998reinforcement, behrens2007learning}. 

\begin{figure}
	\centering
	\includegraphics[width=0.95\textwidth,natwidth=610,natheight=642]{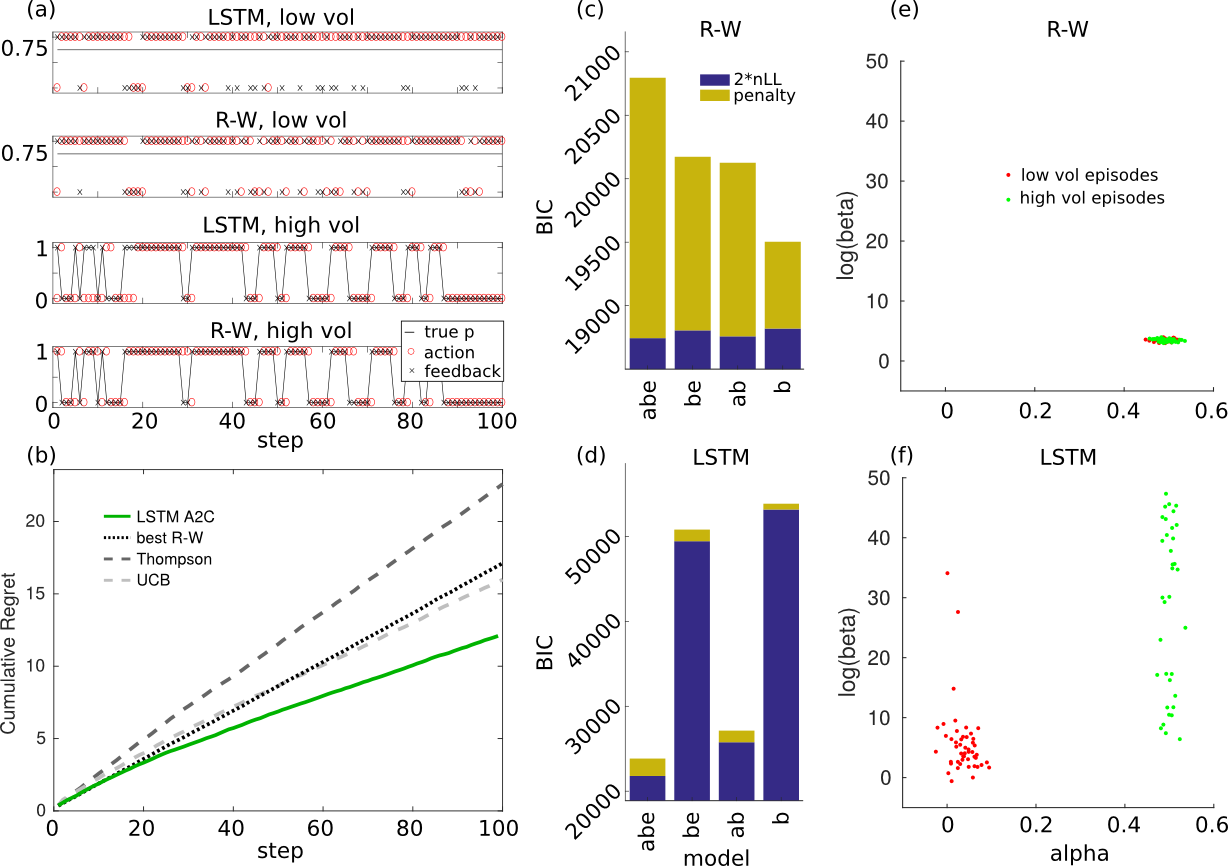}
	\caption{Learned RL procedure adapts its own learning rate to the environment. (a) Agents were trained on two-armed bandits with perfectly anti-correlated Bernoulli reward probabilities, $p_1$ and 1-$p_1$. Two example episodes are shown. $p_1$ changed within an episode (solid black line), with a fast Poisson jump rate in ``high vol'' episodes and a slow rate in ``low vol'' episodes. (b) The trained LSTM agent outperformed UCB, Thompson sampling, and a Rescorla-Wagner (R-W) learner with fixed learning rate $\alpha$=0.5 (selected for being optimal on average in this distribution of environments). (c,d) We fit R-W models by maximum likelihood both to the behavior of R-W (as a control) and to the behavior of LSTM. Models including a learning rate that could vary between episodes (``ab'' and ``abe'') outperformed models without these free parameters on LSTM's data, but not on R-W's data. Addition of a lapse parameter further improved model fits on LSTM's data (``be'' and ``abe''), suggesting that the algorithm implemented by LSTM is not exactly Rescorla-Wagner. (e,f) The LSTM's, but not R-W's, estimated learning rate was higher in volatile episodes. Small jitter added to visualize overlapping points.  }
	\label{fig:volbandits}
\end{figure}

We tested whether meta-RL would learn such a flexible RL policy using a two-armed Bernoulli bandit task with reward probabilities $p_1$ and 1-$p_1$. The value of $p_1$ changed slowly in ``low vol'' episodes and quickly in ``high vol'' episodes. The agent had no way of knowing which type of episode it was in, except for its reward history within the episode. Figure \ref{fig:volbandits}a shows example ``low vol'' and ``high vol'' episodes. Reward magnitude was fixed at 1, and episodes were 100 steps long. UCB and Thompson sampling were again implemented for comparison. The confidence bound term $\sqrt{\frac{\chi \log n}{n_i}}$ in UCB had parameter $\chi$ which was set to 1, selected empirically for good performance on our data set. Thompson sampling's posterior update included knowledge of the Gaussian random walk, but with a fixed volatility for all episodes. 

As in the previous experiment, meta-RL achieved lower regret in test than Thompson sampling, UCB, or the Rescorla-Wagner (R-W) learning rule \citep[Figure \ref{fig:volbandits}b;][]{rescorla1972theory} with the best fixed learning rate ($\alpha$=0.5). To test whether the agent adjusted its effective learning rate to match environments with different volatility levels, we fit R-W models to the agent's behavior, concatenating episodes into blocks of 10, where each block consisted of only ``low vol'' or only ``high vol'' episodes. We considered four different models encompassing different combinations of three parameters: learning rate $\alpha$, softmax inverse temperature $\beta$, and a lapse rate $\epsilon$ to account for unexplained choice variance not related to estimated value \cite{economides2015}. Model ``b'' included only $\beta$, ``ab'' included $\alpha$ and $\beta$, ``be'' included $\beta$ and $\epsilon$, and ``abe'' included all three. All parameters were estimated separately on each block of 10 episodes. In models where $\epsilon$ and $\alpha$ were not free, they were fixed to 0 and 0.5, respectively. Model comparison by Bayesian Information Criterion (BIC) indicated that meta-RL's behavior was better described by a model with different learning rates for each block than a model with a fixed learning rate across blocks. As a control, we performed the same model comparison on the behavior produced by the best R-W agent, finding no benefit of allowing different learning rates across episodes (models ``abe'' and ``ab'' vs ``be'' and ``b''; Figure \ref{fig:volbandits}c-d). In these models, the parameter estimates for meta-RL's behavior were strongly related to the volatility of the episodes, indicating that meta-RL adjusted its learning rate to the volatility of the episode, whereas model fitting the R-W behavior simply recovered the fixed parameters (Figure \ref{fig:volbandits}e-f).

\subsection{Markov decision problems}

The foregoing experiments focused on bandit tasks in which actions do not affect the task's underlying state.  We turn now to MDPs where actions do influence state. We begin with a task derived from the neuroscience literature and then turn to a task, originally studied in the context of animal learning, which requires learning of abstract task structure. As in the previous experiments, our focus is on examining how meta-RL adapts to invariances in task structure. We wrap up by reviewing an experiment recently reported in a related paper \citep{MirowskiICLR17}, which demonstrates how meta-RL can scale to large-scale navigation tasks with rich visual inputs. 

\subsubsection{The ``two-step task''}

Here we examine meta-RL in a setting that has been widely used in the neuroscience literature to distinguish the contribution of different systems viewed to support decision making \citep{daw2005uncertainty}. Specifically, this paradigm -- known as the ``two-step task'' \citep{daw2011model} -- was developed to dissociate a model-free system that caches values of actions in states \citep[e.g. TD(1) Q-learning; see][]{sutton1998reinforcement}, from a model-based system which learns an internal model of the environment and evaluates the value of actions at the time of decision-making through look-ahead planning \citep{daw2005uncertainty}. Our interest was in whether meta-RL would give rise to behavior emulating a model-based strategy, despite the use of a model-free algorithm (in this case A2C) to train the system weights.  

\begin{figure}
    \centering
    \subfloat[\scriptsize{Two-step task}]{
    \includegraphics[width=0.15\textwidth,natwidth=610,natheight=642]{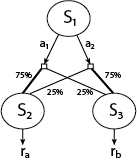}
    }
    \subfloat[\scriptsize{Model predictions}]{
    \includegraphics[width=0.6\textwidth,natwidth=610,natheight=642]{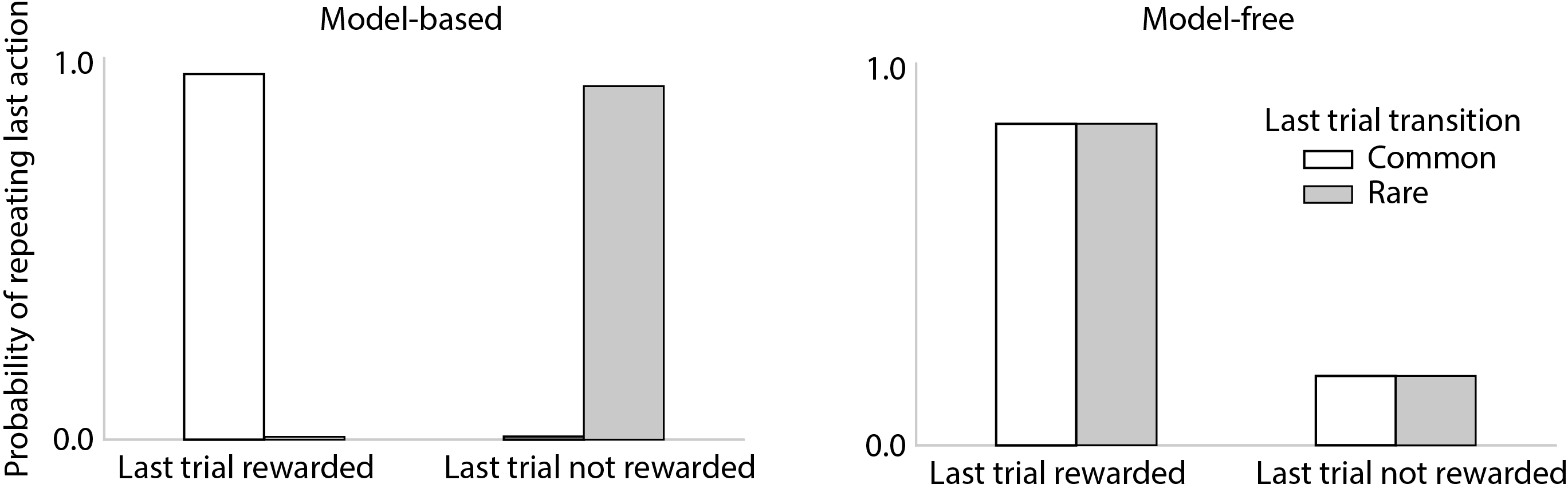}
    }\\
    \subfloat[\scriptsize{LSTM A2C with reward input}]{
    \includegraphics[width=0.35\textwidth,natwidth=610,natheight=642]{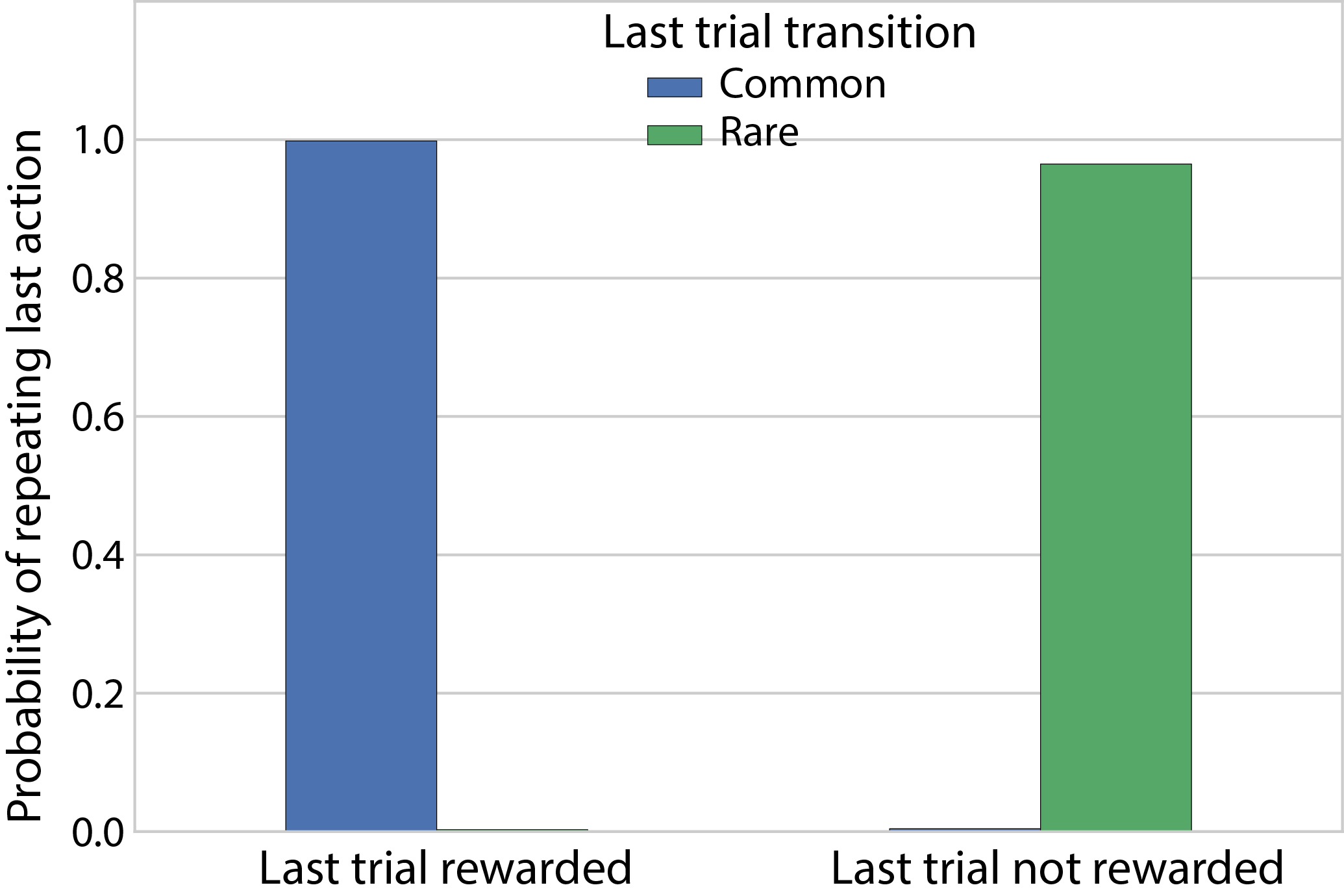}
    }
    \caption{Three-state MDP modeled after the ``two-step task'' from \cite{daw2011model}. (a) MDP with 3 states and 2 actions. All trials start in state $S_1$, with transition probabilities after taking actions $a_1$ or $a_2$ depicted in the graph. $S_2$ and $S_3$ result in expected rewards $r_a$ and $r_b$ (see text). (b) Predictions of choice probabilities given either a model-based strategy or a model-free strategy \citep{daw2011model}. Specifically, model-based strategies take into account transition probabilities and would predict an interaction between the amount of reward received on the last trial and the transition (common or uncommon) observed. (c) Agent displays a perfectly model-based profile when given the reward as input.}
    \label{fig:twostep}
\end{figure}

We used a modified version of the two-step task, designed to bolster the utility of model-based over model-free control  \citep[see][]{kool2016does}. The task's structure is diagrammed in Figure \ref{fig:twostep}a. From the first-stage state $S_1$, action $a_1$ leads to second-stage states $S_2$ and $S_3$ with probability 0.75 and 0.25, respectively, while action $a_2$ leads to $S_2$ and $S_3$ with probabilities 0.25 and 0.75. One second-stage state yielded a reward of 1.0 with probability 0.9 (and otherwise zero); the other yielded the same reward with probability 0.1. The identity of the higher-valued state was assigned randomly for each episode. Thus, the expected values for the two first-stage actions were either $r_a$ = 0.9 and $r_b$ = 0.1, or $r_a$ = 0.1 and $r_b$ = 0.9. All three states were represented by one-hot vectors, with the transition model held constant across episodes: i.e. only the expected value of the second stage states changed from episode to episode. 

We applied the conventional analysis used in the neuroscience literature to dissociate model-free from model-based control  \citep{daw2011model}. This focuses on the ``stay probability,'' that is, the probability with which a first-stage action is selected at trial $t+1$ following a second-stage reward at trial $t$, as a function of whether trial $t$ involved a common transition (e.g. action $a_1$ at state $S_1$ led to $S_2$) or rare transition (action $a_2$ at state $S_1$ led to $S_3$). Under the standard interpretation \citep[see][]{daw2011model}, model-free control -- \textit{{\`a} la} TD(1) -- predicts that there should be a main effect of reward: First-stage actions will tend to be repeated if followed by reward, regardless of transition type, and such actions will tend not to be repeated (choice switch) if followed by non-reward (Figure \ref{fig:twostep}b). In contrast, model-based control predicts an interaction between the reward and transition type, reflecting a more goal-directed strategy, which takes the transition structure into account. Intuitively, if you receive a second-stage reward (e.g. at $S_2$) following a rare transition (i.e. having taken action $a_2$ at state $S_1$), to maximize your chances of getting to this reward on the next trial based on your knowledge of the transition structure, the optimal first stage action is $a_1$ (i.e. switch). 

The results of the stay-probability analysis performed on the agent's choices show a pattern conventionally interpreted as  implying the operation of model-based control (Figure \ref{fig:twostep}c). As in previous experiments, when reward information was withheld at the level of network input, performance was at chance levels. 

If interpreted following standard practice in neuroscience, the behavior of the model in this experiment reflects a surprising effect: training with model-free RL gives rise to behavior reflecting model-based control. We hasten to note that different interpretations of the observed pattern of behavior are available \citep{akam2015simple}, a point to which we will return below. However, notwithstanding this caveat, the results of the present experiment provide a further illustration of the point that the learning procedure that emerges from meta-RL can differ starkly from the original RL algorithm used to train the network weights, and takes a form that exploits consistent task structure. 

\subsubsection{Learning abstract task structure}

In the final experiment we conducted, we took a step towards examining the scalabilty of meta-RL, by studying a task that involves rich visual inputs, longer time horizons and sparse rewards. Additionally, in this experiment we studied a meta-learning task that requires the system to tune into an abstract task structure, in which a series of objects play defined roles which the system must infer. 

The task was adapted from a classic study of animal behavior, conducted by \cite{harlow1949formation}. On each trial in the original task, Harlow presented a monkey with two visually contrasting objects. One of these covered a small well containing a morsel of food; the other covered an empty well. The animal chose freely between the two objects and could retrieve the food reward if present. The stage was then hidden and the left-right positions of the objects were randomly reset. A new trial then began, with the animal again choosing freely. This process continued for a set number of trials using the same two objects. At completion of this set of trials, two entirely new and unfamiliar objects were substituted for the original two, and the process began again. Importantly, within each block of trials, one object was chosen to be consistently rewarded (regardless of its left-right position), with the other being consistently unrewarded. What Harlow \citep{harlow1949formation} observed was that, after substantial practice, monkeys displayed behavior that reflected an understanding of the task's rules. When two new objects were presented, the monkey's first choice between them was necessarily arbitrary. But after observing the outcome of this first choice, the monkey was at ceiling thereafter, always choosing the rewarded object. 

\begin{figure}
    \centering
    \subfloat[\scriptsize{Fixation}]{
    \includegraphics[width=0.25\textwidth,natwidth=610,natheight=642]{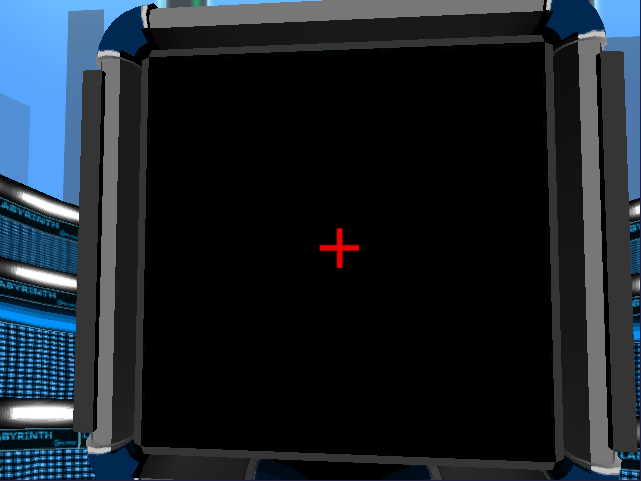}
    }
    \subfloat[\scriptsize{Image display}]{
    \includegraphics[width=0.25\textwidth,natwidth=610,natheight=642]{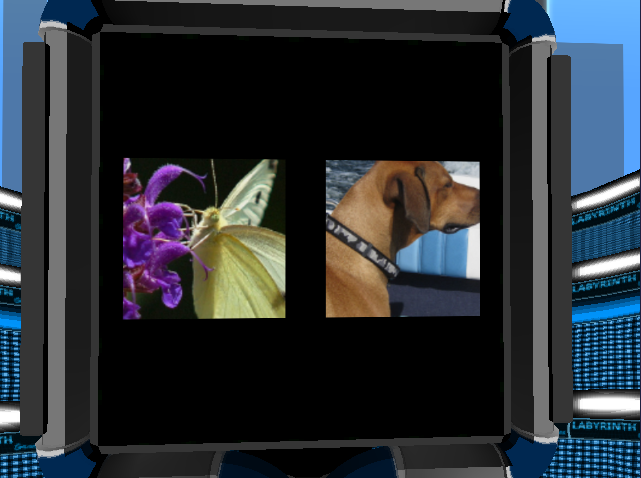}
    }
    \subfloat[\scriptsize{Right saccade and selection}]{
    \includegraphics[width=0.25\textwidth,natwidth=610,natheight=642]{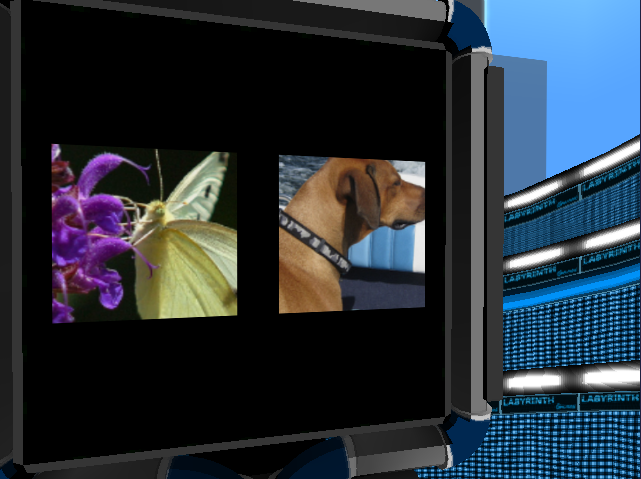}
    }\\
    \subfloat[\scriptsize{Training performance}]{
    \includegraphics[width=0.35\textwidth,natwidth=610,natheight=642]{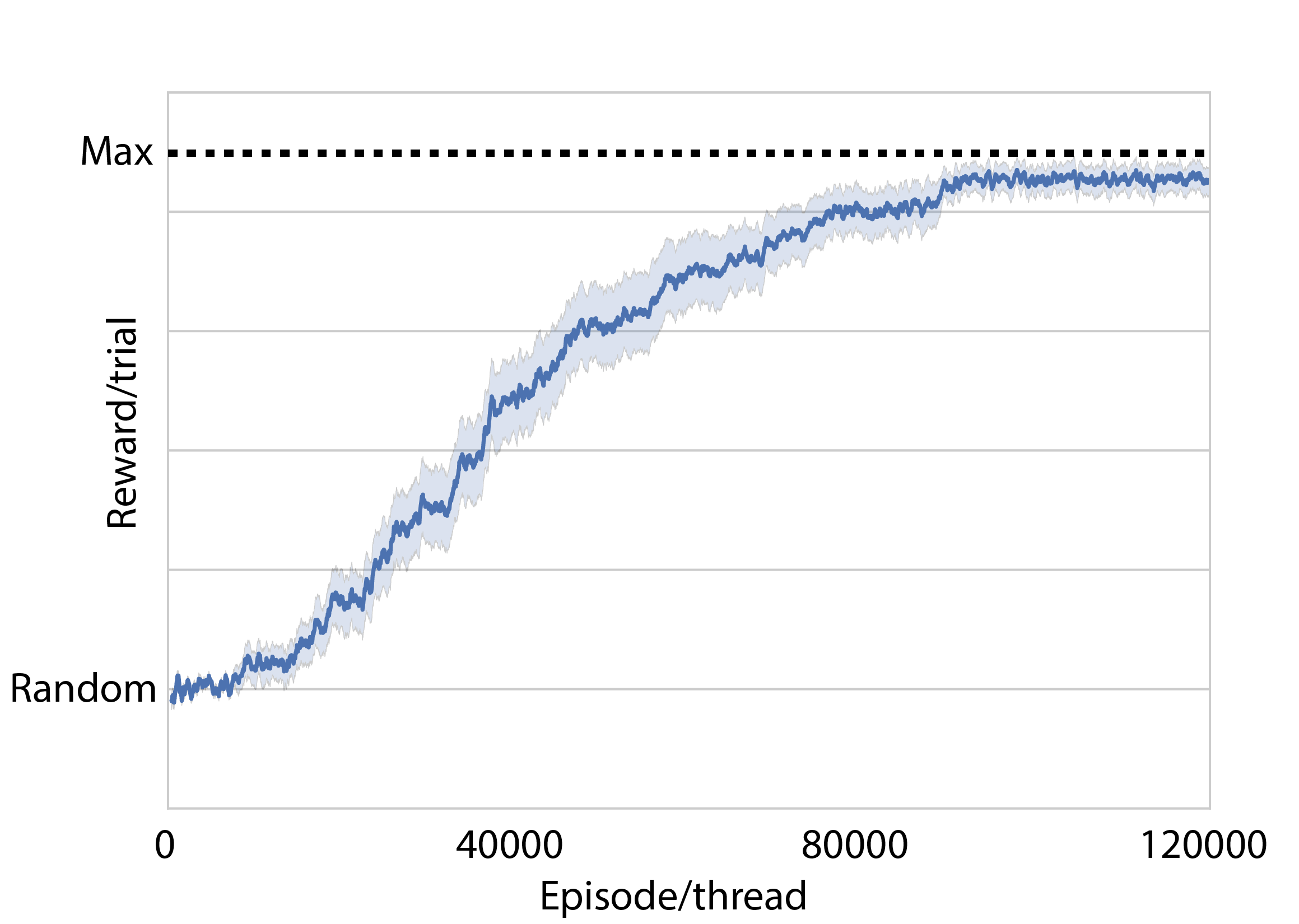}
    }
    \subfloat[\scriptsize{Robustness over random seeds}]{
    \includegraphics[width=0.32\textwidth,natwidth=610,natheight=642]{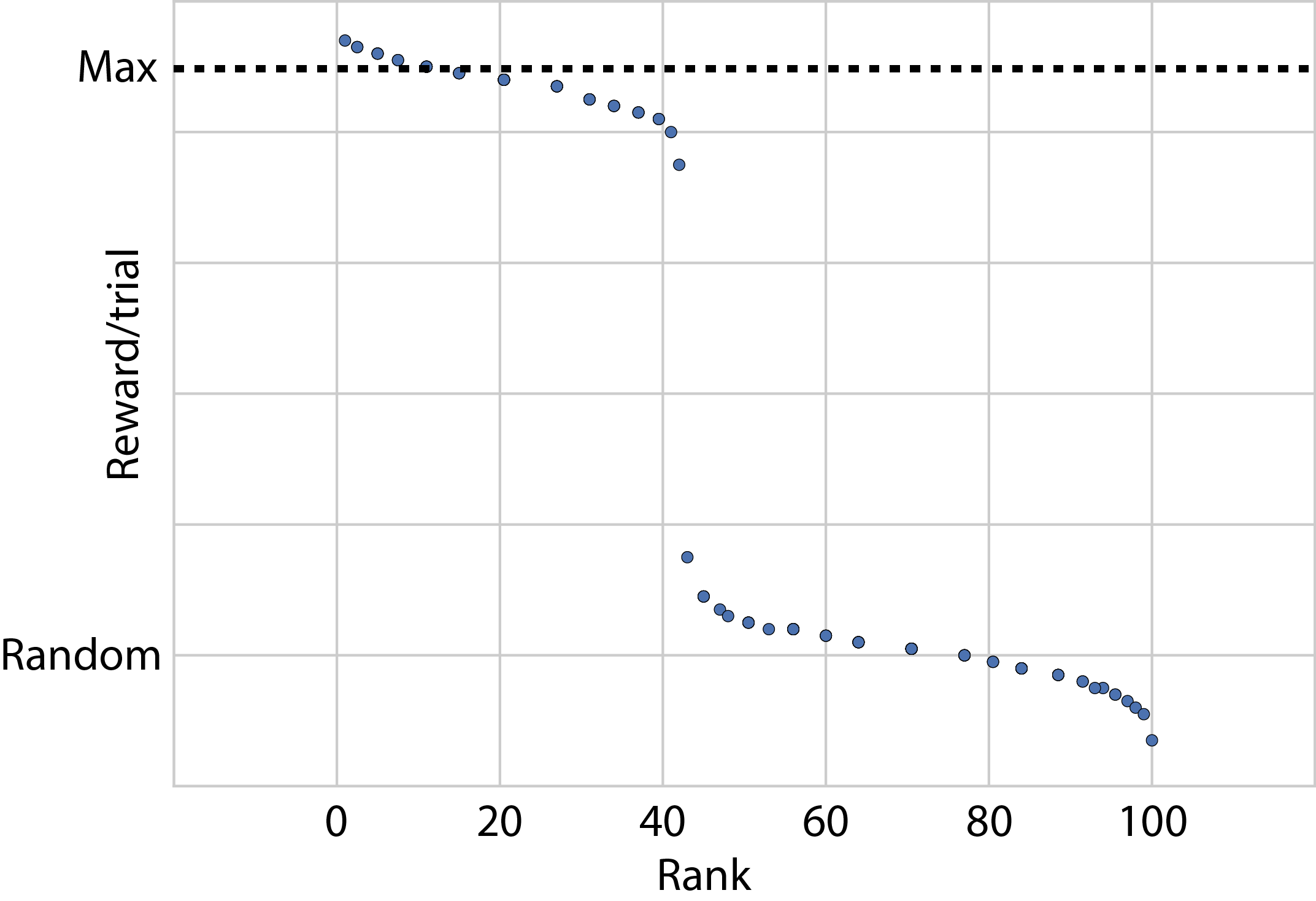}
    }
    \subfloat[\scriptsize{One-shot learning}]{
    \includegraphics[width=0.27\textwidth,natwidth=610,natheight=642]{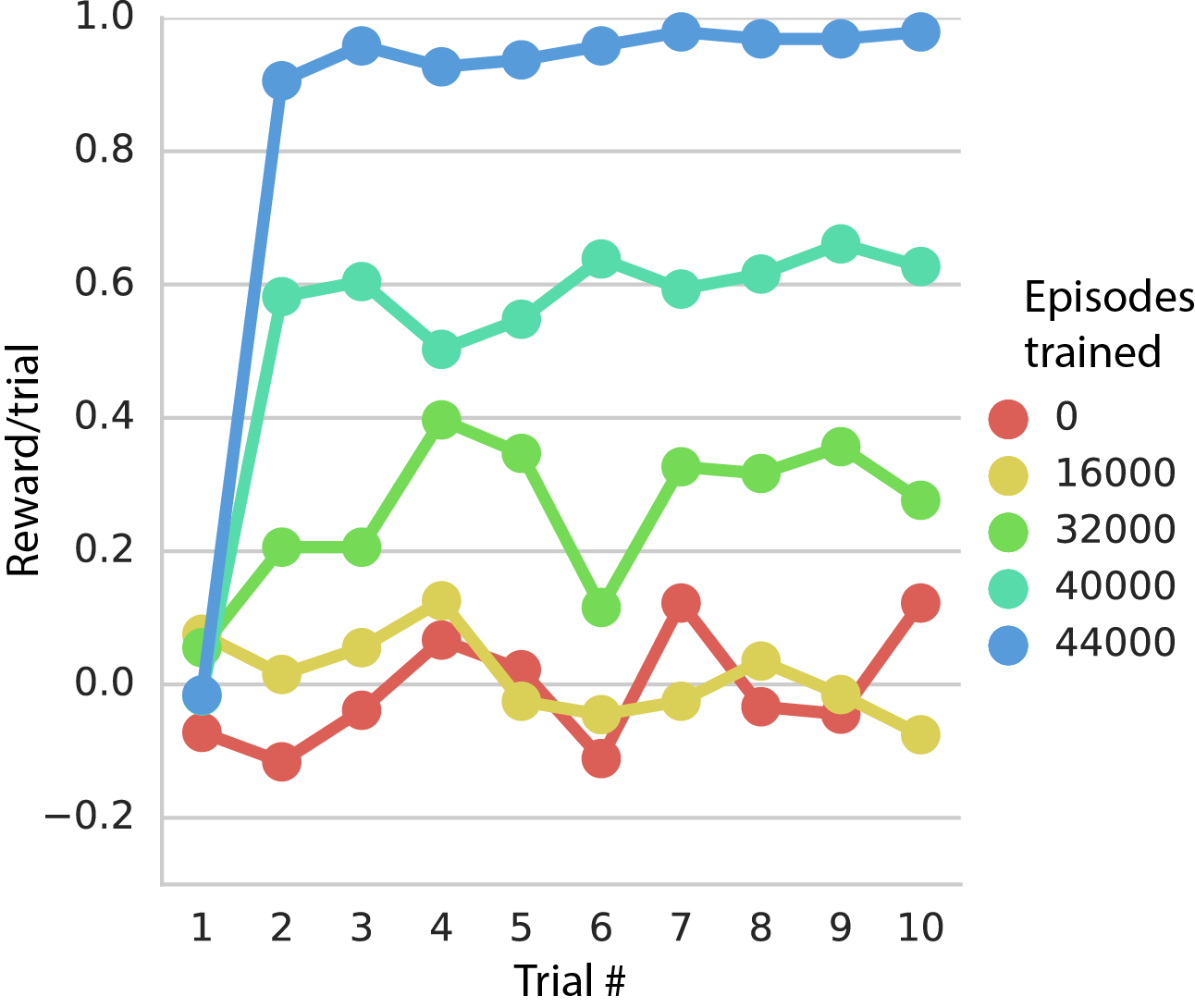}
    }
    \caption{Learning abstract task structure in visually rich 3D environment. a-c) Example of a single trial, beginning with a central fixation, followed by two images with random left-right placement. d) Average performance (measured in average reward per trial) of top 40 out of 100 seeds during training. Maximum expected performance is indicated with black dashed line. e) Performance at episode 100,000 for 100 random seeds, in decreasing order of performance. f) Probability of selecting the rewarded image, as a function of trial number for a single A3C stacked LSTM agent for a range of training durations (episodes per thread, 32 threads).}
    \label{fig:harlow}
\end{figure}

We anticipated that meta-RL should give rise to the same pattern of abstract one-shot learning. In order to test this, we adapted Harlow's paradigm into a visual fixation task, as follows. A 84x84 pixel input represented a simulated computer screen (see Figure \ref{fig:harlow}a-c). At the beginning of each trial, this display was blank except for a small central fixation cross (red crosshairs). The agent selected discrete left-right actions  which shifted its view approximately 4.4 degrees in the corresponding direction, with a small momentum effect (alternatively, a no-op action could be selected). The completion of a trial required performing two tasks: saccading to the central fixation cross, followed by saccading to the correct image. If the agent held the fixation cross  in the center of the field of view (within a tolerance of 3.5 degrees visual angle) for a minimum of four time steps, it received a reward of 0.2. The fixation cross then disappeared and two images -- drawn randomly from the ImageNet dataset \citep{deng2009imagenet} and resized to 34x34 -- appeared on the left and right side of the display (Figure \ref{fig:harlow}b). The agent's task was then to ``select'' one of the images by rotating until the center of the image aligned with the center of the visual field of view (within a tolerance of 7 degrees visual angle). Once one of the images was selected, both images disappeared and, after an intertrial interval of 10 time-steps, the fixation cross reappeared, initiating the next trial. Each episode contained a maximum of 10 trials or 3600 steps. Following \cite{MirowskiICLR17}, we implemented an action repeat of 4, meaning that selecting an image took a minimum of three independent decisions (twelve primitive actions) after having completed the fixation. It should be noted, however, that the rotational position of the agent was not limited; that is, 360 degree rotations could occur, while the simulated computer screen only subtended 65 degrees. 

Although new ImageNet images were chosen at the beginning of each episode (sampled with replacement from a set of 1000 images), the same images were re-used across all trials within an episode, though in randomly varying left-right placement, similar to the objects in Harlow's experiment. And as in that experiment, one image was arbitrarily chosen to be the ``rewarded'' image throughout the episode. Selection of this image yielded a reward of 1.0, while the other image yielded a reward of -1.0. During test, the A3C learning rate was set to zero and ImageNet images were drawn from a separate held-out set of 1000, never presented during training. 

A grid search was conducted for optimal hyperparameters. At perfect performance, agents can complete one trial per 20-30 steps and achieve a maximum expected reward of 9 per 10 trials. Given the nature of the task -- which requires one-shot image-reward memory together with maintenance of this information over a relatively long timescale (i.e. over fixation-cross selections and across trials) -- we assessed the performance of not only a convolutional-LSTM architecture which receives reward and action as additional input (see Figure \ref{fig:archs}b and Table \ref{table:params}), but also a convolutional-stacked LSTM architecture used in a navigation task discussed below (see Figure \ref{fig:archs}c).  

Agent performance is illustrated in Figure \ref{fig:harlow}d-f. Whilst the single LSTM agent was relatively successful at solving the task, the stacked-LSTM variant exhibited much better robustness. That is, 43\% of random seeds of the best hyperparameter set performed at ceiling (Figure \ref{fig:harlow}e), compared to 26\% of the single LSTM.

Like the monkeys in Harlow's experiment \citep{harlow1949formation}, the networks converge on an optimal policy: Not only does the agent successfully fixate to begin each trial, but starting on the second trial of each episode it invariably selects the rewarded image, regardless of which image it selected on the first trial(Figure \ref{fig:harlow}f). This reflects an impressive form of one-shot learning, which reflects an implicit understanding of the task structure: After observing one trial outcome, the agent binds a complex, unfamiliar image to a specific task role. 
 
Further experiments, reported elsewhere \citep{wangSUBmeta}, confirmed that the same recurrent A3C system is also able to solve a substantially more difficult version of the task. In this task, only one image -- which was randomly designated to be either the rewarding item to be selected, or the unrewarding item to be avoided -- was presented on every trial during an episode, with the other image presented being novel on every trial.

\subsubsection{One-shot navigation}
The experiments using the Harlow task demonstrate the capacity of meta-RL to operate effectively within a visually rich environment, with relatively long time horizons. Here we consider related experiments recently reported within the navigation domain \citep{MirowskiICLR17} \citep[see also][]{JaderbergDreamingICLR}, and discuss how these can be recast as examples of meta-RL -- attesting to the scaleability of this principle to more typical MDP settings that pose challenging RL problems due to dynamically changing sparse rewards. 

\begin{figure}
    \centering
    \subfloat[\scriptsize{Labryinth I-maze}]{
    \includegraphics[width=0.4\textwidth,natwidth=610,natheight=642]{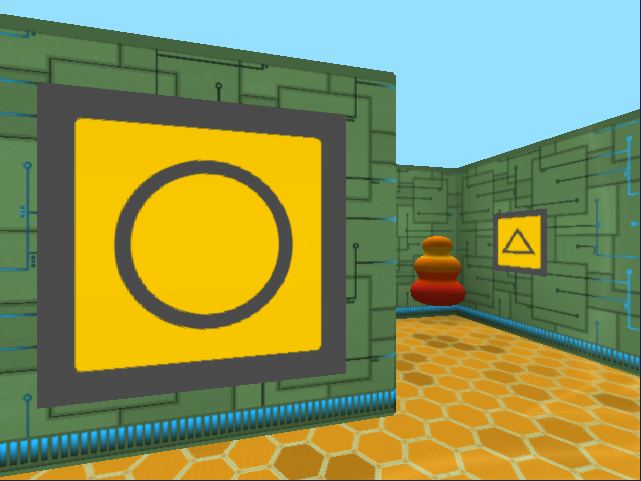}
    }
    \subfloat[\scriptsize{Illustrative Episode}]{
    \includegraphics[width=0.4\textwidth,natwidth=610,natheight=642]{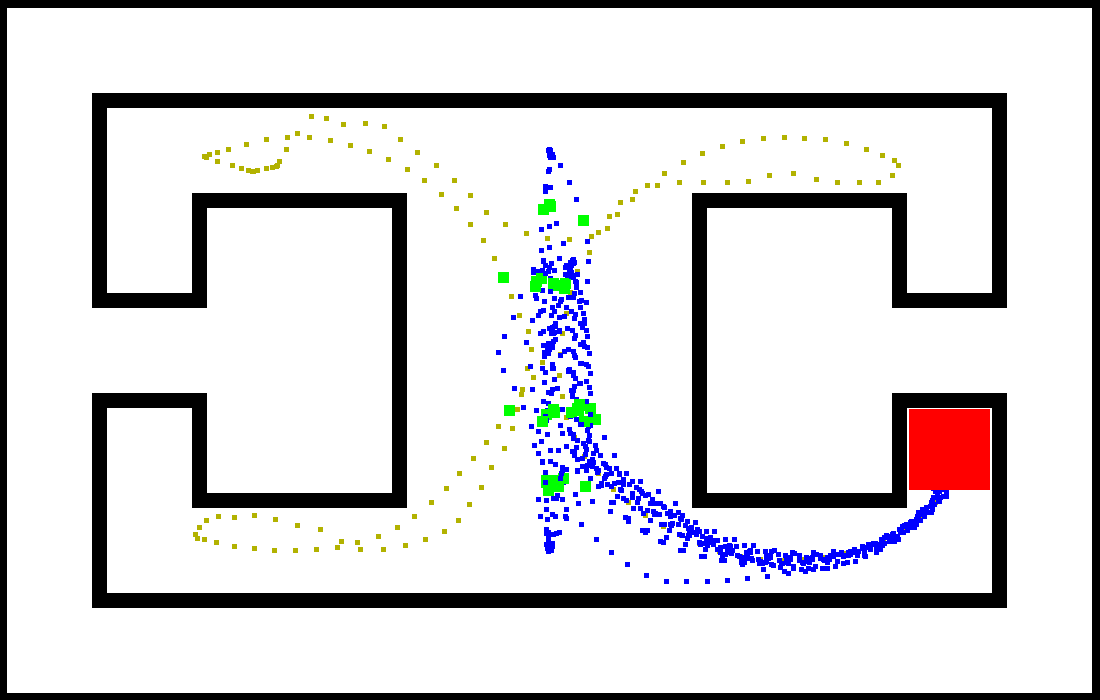}
    }\\
    \subfloat[\scriptsize{Performance}]{
    \includegraphics[width=0.4\textwidth,natwidth=610,natheight=642]{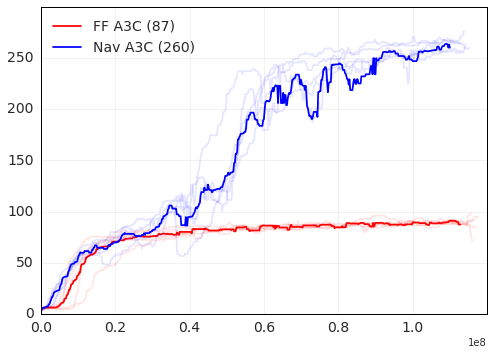}
    }
    \subfloat[\scriptsize{Value Function}]{
    \includegraphics[width=0.4\textwidth,natwidth=610,natheight=642]{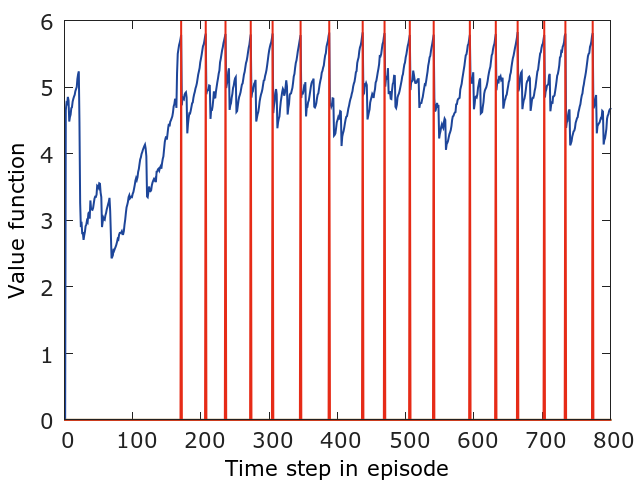}
    }
    \caption{a) view of I-maze showing goal object in one of the 4 alcoves b) following initial exploration (light trajectories), agent repeatedly goes to goal (blue trajectories) c) Performance of stacked LSTM (termed ``Nav A3C'') and feedforward (``FF A3C'') architectures, per episode (goal = 10 points) averaged across top 5 hyperparameters. e) following initial goal discovery (goal hits marked in red), value function occurs well in advance of the agent seeing the goal which is hidden in an alcove. Figure used with permission from \cite{MirowskiICLR17}.}
    \label{fig:imaze}
\end{figure}

Specifically, we consider a setting where the environment layout is fixed but the goal changes location randomly each episode \citep[Figure \ref{fig:imaze};][]{MirowskiICLR17}. Although the layout is relatively simple, the Labyrinth environment \citep[see for details][]{MirowskiICLR17} is richer and more finely discretized (cf VizDoom), resulting in long time horizons; a trained agent takes approximately 100 steps (10 seconds) to reach the goal for the first time in a given episode. Results show that a stacked LSTM architecture (Figure \ref{fig:archs}c), that receives reward and action as additional inputs equivalent to that used in our Harlow experiment achieves near-optimal behavior -- showing one-shot memory for the goal location after an initial exploratory period, followed by repeated exploitation (see Figure \ref{fig:imaze}c). This is evidenced by a substantial decrease in latency to reach the goal for the first time (\textasciitilde100 timesteps) compared to subsequent visits (\textasciitilde30 timesteps). Notably, a feedforward network (see Figure \ref{fig:imaze}c), that receives only a single image as observation, is unable to solve the task (i.e. no decrease in latency between successive goal rewards).  Whilst not interpreted as such in \cite{MirowskiICLR17}, this provides a clear demonstration of the effectiveness of meta-RL: a separate RL algorithm with the capability of one-shot learning emerges through training with a fixed and more incremental RL algorithm (i.e. policy gradient). Meta-RL can be viewed as allowing the agent to infer the optimal value function following initial exploration (see Figure \ref{fig:imaze}d) -- with the additional LSTM providing information about the currently relevant goal location to the LSTM that outputs the policy over the extended timeframe of the episode. Taken together, meta-RL allows a base model-free RL algorithm to solve a challenging RL problem that might otherwise require fundamentally different approaches (e.g. based on successor representations or fully model-based RL).

\section{Related work}

We have already touched on the relationship between deep meta-RL and pioneering work by \cite{hochreiter2001learning} using recurrent networks to perform meta-learning in the setting of full supervision \citep[see also][]{prokhorov2002adaptive,cotter1990fixed,younger1999fixed}. That approach was recently extended in \cite{santoro2016meta}, which demonstrated the utility of leveraging an external memory structure. The idea of crossing meta-learning with reinforcement learning has been previously discussed by \cite{schmidhuber1996simple}. That work, which appears to have introduced the term ``meta-RL,'' differs from ours in that it did not involve a neural network implementation. More recently, however, there has been a surge of interest in using neural networks to learn optimization procedures, using a range of innovative meta-learning techniques \citep{li2016learning,zoph2016neural,andrychowicz2016learning,chen2016blackbox}. Recent work by \cite{chen2016blackbox} is particularly close in spirit to the work we have presented here, and can be viewed as treating the case of ``infinite bandits'' using a meta-learning strategy broadly analogous to the one we have pursued.

The present research also bears a close relationship with a different body of recent work that has not been framed in terms of meta-learning. A number of studies have used deep RL to train recurrent neural networks on navigation tasks, where the structure of the task (e.g., goal location or maze configuration) varies across episodes \citep{MirowskiICLR17, JaderbergDreamingICLR}. The final experiment that we presented above, drawn from \citep{MirowskiICLR17}, is one example.  To the extent that such experiments involve the key ingredients of deep meta-RL -- a neural network with memory, trained through RL on a series of interrelated tasks -- they are almost certain to involve the kind of meta-learning we have described in the present work. This related work provides an indication that meta-RL can be fruitfully applied to larger scale problems than the ones we have studied in our own experiments. Importantly, it indicates that a key ingredient in scaling the approach may be to incorporate memory mechanisms beyond those inherent in unstructured recurrent neural networks \citep[see][]{MirowskiICLR17,santoro2016meta,graves2016hybrid,weston2014memory}. Our work, for its part, suggests that there is untapped potential in deep recurrent RL agents to meta-learn quite abstract aspects of task structure, and to discover strategies that exploit such structure toward rapid, flexible adaptation. 

During completion of the present research, closely related work was reported by \cite{duan2016RL2}. Like us, Duan and colleagues use deep RL to train a recurrent network on a series of interrelated tasks, with the result that the network dynamics learn a second RL procedure which operates on a faster time-scale than the original algorithm. They compare the performance of these learned procedures against conventional RL algorithms in a number of domains, including bandits and navigation. An important difference between this parallel work and our own is the former's primary focus on relatively unstructured task distributions (e.g., uniformly distributed bandit problems and random MDPs); our main interest, in contrast, has been in structured task distributions \citep[e.g., dependent bandits and the task introduced by][]{harlow1949formation}, because it is in this setting where the system can learn a biased -- and therefore efficient -- RL procedure that exploits regular task structure. The two perspectives are, in this regard, nicely complementary. 

\section{Conclusion}
A current challenge in artificial intelligence is to design agents that can adapt rapidly to new tasks by leveraging knowledge acquired through previous experience with related activities.  In the present work we have reported initial explorations of what we believe is one promising avenue toward this goal. Deep meta-RL involves a combination of three ingredients: (1) Use of a deep RL algorithm to train a recurrent neural network, (2) a training set that includes a series of interrelated tasks, (3) network input that includes the action selected and reward received in the previous time interval. The key result, which emerges naturally from the setup rather than being specially engineered, is that the recurrent network dynamics learn to implement a second RL procedure, independent from and potentially very different from the algorithm used to train the network weights.  Critically, this learned RL algorithm is tuned to the shared structure of the training tasks. In this sense, the learned algorithm builds in domain-appropriate biases, which can allow it to operate with greater efficiency than a general-purpose algorithm. This bias effect was particularly evident in the results of our experiments involving dependent bandits (sections 3.1.2 and 3.1.3), where the system learned to take advantage of the task's covariance structure; and in our study of Harlow's animal learning task (section 3.2.2), where the recurrent network learned to exploit the task's structure in order to display one-shot learning with complex novel stimuli. 

One of our experiments (section 3.2.1) illustrated the point that a system trained using a model-free RL algorithm can develop behavior that emulates model-based control. A few further comments on this result are warranted. As noted in our presentation of the simulation results, the pattern of choice behavior displayed by the network has been considered in the cognitive and neuroscience literatures as reflecting model-based control or tree search. However, as has been remarked in very recent work, the same pattern can arise from a model-free system with an appropriate state representation \citep{akam2015simple}. Indeed, we suspect this may be how our network in fact operates. However, other findings suggest that a more explicitly model-based control mechanism can emerge when a similar system is trained on a more diverse set of tasks. In particular, \cite{ilin2007efficient} showed that recurrent networks trained on random mazes can approximate dynamic programming procedures \citep[see also][]{silver2016predictron,abbeel2016vin}. At the same time, as we have stressed, we consider it an important aspect of deep meta-RL that it yields a learned RL algorithm that capitalizes on invariances in task structure. As a result, when faced with widely varying but still structured environments, deep meta-RL seems likely to generate RL procedures that occupy a grey area between model-free and model-based RL.

The two-step decision problem studied in Section 3.2.1 was derived from neuroscience, and we believe deep meta-RL may have important implications in that arena \citep{wangSUBmeta}. The notion of meta-RL has been discussed previously in neuroscience but only in a narrow sense, according to which meta-learning adjusts scalar hyperparameters such as the learning rate or softmax inverse temperature \citep{lee2009mechanisms,schweighofer2003meta,soltani2006neural,khamassi2011robot,kobayashi2009meta,khamassi2013medial}. In recent work \citep{wangSUBmeta} we have shown that deep meta-RL can account for a wider range of experimental observations, providing an integrative framework for understanding the respective roles of dopamine and the prefrontal cortex in biological reinforcement learning.

ACKNOWLEDGEMENTS

We would like the thank the following colleagues for useful discussion and feedback: Nando de Freitas, David Silver, Koray Kavukcuoglu, Daan Wierstra, Demis Hassabis, Matt Hoffman, Piotr Mirowski, Andrea Banino, Sam Ritter, Neil Rabinowitz, Peter Dayan, Peter Battaglia, Alex Lerchner, Tim Lillicrap and Greg Wayne. 

\footnotesize
\setlength{\bibsep}{5pt}
\bibliographystyle{plainnat}
\bibliography{metaRL}

\newpage




\end{document}